\documentclass[conference]{IEEEtran}
\usepackage{times}

\usepackage[numbers]{natbib}
\usepackage{multicol}
\usepackage{graphicx}
\usepackage{booktabs}
\usepackage{multirow}
\usepackage{amsmath,amssymb}
\usepackage{tikz}
\usetikzlibrary{positioning, arrows.meta}
\usepackage[bookmarks=true]{hyperref}
\graphicspath{{./}{figures/}}

\begin{document}

\title{\LARGE Changepoint-Aware World Models: Detecting Dynamics Shifts and\\Recovering by Forgetting Stale Replay in Model-Based RL}

\author{\authorblockN{\large Everest Yang}
\authorblockA{\large Brown University\\
everest\_yang@brown.edu}}

\maketitle
\thispagestyle{empty}
\pagestyle{empty}

\begin{abstract}
A robot's learned model of its own dynamics is only valid until those dynamics
change: actuators wear, payloads shift, and joints stiffen. A model-based agent that
keeps training as if nothing happened adapts slowly, dragged back by a replay buffer
full of stale experience. We present \emph{Changepoint-Aware World Models} (CAWM), a
DreamerV3 agent that detects an abrupt dynamics shift from its \emph{own} internal
prediction error, using an online CUSUM test against a rolling baseline that fires only
on abrupt change rather than on slow learning drift. It then \emph{forgets stale
replay}, keeping the learned representation while flushing obsolete data. On simulated
locomotion under two robot-relevant shifts, doubled gravity and halved actuator gain,
CAWM recovers substantially faster than passive retraining. It also beats a strong
baseline that respawns a fresh dynamics model on detection, the deep-world-model
analogue of model-bank methods. With the response triggered at the shift, CAWM gains
$+95$ to $+153$ return in the first $30$k post-shift frames over three seeds, while
matching that respawn at asymptote. Running the detector in closed loop reproduces this
gain on the gravity shift. The benefit holds across both shift types, and is largest
when the shift is severe enough that old data is genuinely obsolete.
\end{abstract}

\IEEEpeerreviewmaketitle

\begin{figure*}[t]
\centering
\begin{tikzpicture}[
    font=\small, node distance=7mm and 12mm,
    box/.style={draw, rounded corners=2pt, align=center, inner sep=5pt,
                minimum height=11mm, minimum width=26mm},
    accent/.style={box, fill=blue!12, draw=blue!65!black},
    plain/.style={box, fill=black!5, draw=black!45},
    arr/.style={-{Stealth[length=2mm]}, thick, black!75},
]
\node[plain] (env) {Environment\\ (abrupt shift)};
\node[plain, right=of env] (wm) {DreamerV3 world\\ model (RSSM)};
\node[plain, right=of wm] (sig) {internal signal\\ $\mathcal{L}_{\mathrm{dyn}}(t)$};
\node[accent, right=of sig] (cusum) {online CUSUM\\ $S_t\!=\!\max(0,S_{t\text{-}1}\!+\!z_t\!-\!k)$};
\node[plain, below=of cusum] (det) {changepoint\\ $S_t\!>\!h \Rightarrow \hat\tau$};
\node[accent, left=of det] (forget) {forget stale replay\\ $|\mathcal{D}|\!\leftarrow\! N_{\mathrm{keep}}$};
\node[plain, left=of forget] (refit) {re-fit on\\ fresh data};
\node[plain, left=of refit] (rec) {fast recovery};
\draw[arr] (env) -- (wm); \draw[arr] (wm) -- (sig); \draw[arr] (sig) -- (cusum);
\draw[arr] (cusum) -- (det); \draw[arr] (det) -- (forget); \draw[arr] (forget) -- (refit);
\draw[arr] (refit) -- (rec);
\draw[arr] (rec.west) -| ([xshift=-13mm]rec.west) |- ([yshift=7mm]env.north) -| (wm.north);
\end{tikzpicture}
\caption{Changepoint-Aware World Model. The agent monitors its own world-model
prediction error $\mathcal{L}_{\mathrm{dyn}}$. An online CUSUM test against a rolling
trailing baseline fires only on an abrupt dynamics shift (blue: the changepoint
machinery). On detection it forgets stale replay, capping the buffer while keeping the
encoder, so the world model re-fits the new dynamics and the agent recovers. The
MBCD-style baseline instead respawns the dynamics model at $\hat\tau$.}
\label{fig:pipeline}
\end{figure*}
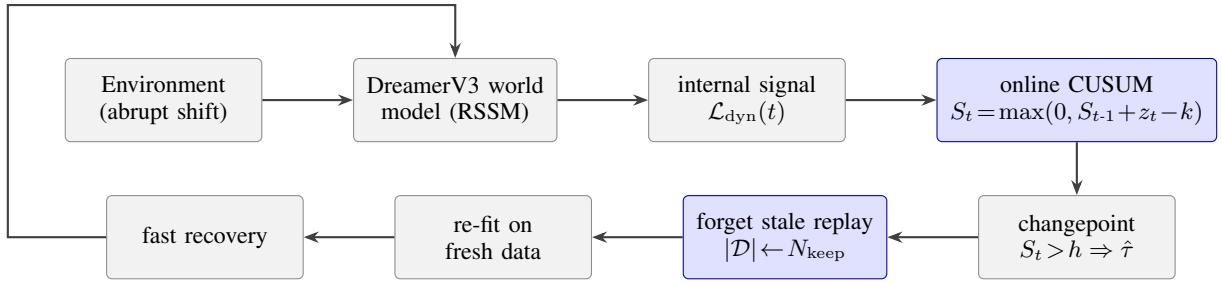

\section{Introduction}
A model-based reinforcement learning agent builds an internal model of how its actions
move the world, and then plans or imagines inside it. For a robot, that is a model of
its own body and contacts. But a real robot's dynamics are not stationary: actuators
degrade, payloads are picked up, joints stiffen with wear, terrain changes underfoot.
When the dynamics shift abruptly, a world model trained on the old regime is suddenly
wrong, and an agent that keeps training recovers slowly, held back by a replay buffer
dominated by stale transitions.

This raises two coupled questions. First, can an agent \emph{detect} that its dynamics
have shifted using only signals it already computes, without an external change
oracle? Second, once it knows, what is the right \emph{response}? We study both, and
separate them: detection first, then the response on top.

Our \textbf{primary contribution} is that in a deep latent world model, the most
effective response to a detected shift is to \emph{forget stale experience} while
\emph{keeping the learned representation}. We
detect the shift with an online CUSUM test \cite{cusum} on the world model's own
dynamics-prediction loss, against a \emph{rolling trailing baseline} that tracks the
agent's slow learning drift so that only abrupt changes trigger it. On detection we
cap the replay buffer, evicting obsolete data and let the world model re-fit. The
encoder and policy are untouched and only the data the model trains on is refreshed.

Our \textbf{secondary contribution} is a controlled comparison against the natural
alternative: respawning a fresh dynamics model on detection, the deep-world-model
analogue of model-bank changepoint methods \cite{mbcd}. We find that respawning incurs
a sharp re-adaptation transient, a second crash as the fresh model relearns from
scratch, whereas forgetting avoids it. The two reach comparable asymptotic performance,
so the advantage of CAWM is \emph{faster recovery}. This holds across two robot-relevant shift types and three seeds each.

In summary, we contribute: (i) an online changepoint detector that reads a deep world
model's internal prediction error against a rolling baseline. (ii) A forgetting
response that recovers faster than passive retraining and faster than respawning a
fresh model, at comparable asymptote, across a gravity shift and an actuator-wear
shift. (iii) A conditional finding delimiting when forgetting helps, namely when the
shift is severe enough that old data is obsolete.

\section{Related Work}

\textbf{World models and model-based RL:}
Learning a model of the environment and planning or training a policy inside it is a
long-standing idea \cite{worldmodels, dreamerv1, muzero, pets, tdmpc}. DreamerV3
\cite{dreamerv3}, building on DreamerV2 \cite{dreamerv2}, learns a latent world model
and trains an actor-critic inside it, and is our base agent. Such models, and the
actor-critics they build on \cite{sac}, are typically studied under stationary dynamics.
Exploration methods that act on model disagreement \cite{plan2explore} detect where the
model is uncertain, but target a fixed environment, not a changing one.

\textbf{Changepoint detection and adaptation in RL:}
MBCD \cite{mbcd} runs a multivariate CUSUM over an ensemble of dynamics predictors and
switches a context model on detection. Our respawn baseline is its analogue inside a
single deep world model. Closest to our detector, \citet{noveltywm} flag a shift from a
DreamerV2 world model's prediction misalignment, but for \emph{detection only}, leaving
recovery out of scope. We instead detect \emph{and} respond. ARCADE \cite{arcade} tempers
a Bayesian dynamics prior on a changepoint, the closest neighbour to our forgetting, but
freezes an offline representation and adapts a linear decoder rather than learning
end-to-end. Other non-stationary methods instead adapt the policy, by detecting a
behaviour change \cite{bada}, inferring a hidden latent context \cite{secbad, hipmdp,
lilac}, or meta-learning across dynamics \cite{continuousadapt}. More broadly, continual
RL studies the stability-plasticity trade-off our forgetting navigates \cite{continualrl,
abel}, including the loss of plasticity stale data induces \cite{ewc, primacy, plasticity}.

\textbf{Robot adaptation:}
Rapid Motor Adaptation \cite{rma} adapts a legged policy to payload, wear, and terrain
by inferring a latent context online. We share the motivation, but address it inside a
world model by detecting the change and refreshing its data, rather than context
inference in a model-free policy.

\section{Method}
\textbf{Problem:} An agent acts in an environment whose transition dynamics
$P_\theta$ are stationary until an unknown changepoint $\tau$, after which they switch
abruptly to $P_{\theta'}$ and remain fixed. The agent observes only its own states,
actions, and rewards. The changepoint $\tau$ and the new dynamics are unknown. We measure post-shift
performance by episode return over fixed windows relative to $\tau$: a
\emph{transient} window $[0,30\mathrm{k})$ frames, an \emph{asymptote} window
$[30\mathrm{k},90\mathrm{k})$, and their union, the \emph{cumulative} window, which is
a return-based proxy for adaptation regret.

\textbf{Detecting the shift:} A DreamerV3 world model already computes a
dynamics-prediction loss $\mathcal{L}_{\mathrm{dyn}}(t)$ (the KL between the posterior
and prior latent transition) each update. When the true dynamics shift, the model's
predictions stop matching observation and $\mathcal{L}_{\mathrm{dyn}}$ rises. The
difficulty is that $\mathcal{L}_{\mathrm{dyn}}$ is also non-stationary \emph{before}
any shift, as the agent is still learning, so a fixed-threshold test false-alarms on
the learning curve itself. We therefore run a one-sided CUSUM test \cite{cusum, bocpd}
against a \emph{rolling trailing baseline}. We standardize $\mathcal{L}_{\mathrm{dyn}}(t)$
by the mean and standard deviation of a trailing window, lagged to exclude the current
point. The statistic $S_t=\max(0,\,S_{t-1}+z_t-k)$ then accumulates only sustained positive deviations and fires when $S_t>h$. Because the baseline slides with the agent's slow drift, gradual learning is absorbed and only an abrupt jump crosses the threshold. The slack $k$ and threshold $h$ trade detection delay against false alarms, and we use a fixed $k{=}0.5$ and $h{=}8$. We monitor the dynamics loss rather than a one-step latent surprise computed on fresh transitions which we found too weak and noisy to trigger reliably. Both are internal, control-relevant signals the model produces in the spirit of the value-equivalence view of what a model is for \cite{valueequiv}.

\textbf{Responding by forgetting:} On detection, CAWM caps the replay buffer to its
most recent $N_{\mathrm{keep}}$ transitions. DreamerV3 already evicts the oldest
episodes down to a configured size. A single change to that size flushes stale pre-shift
data within roughly $N_{\mathrm{keep}}$ frames, after which the world model trains only
on post-shift experience. The encoder, actor, and critic are left intact, and only the data is refreshed. This differs from prioritized experience replay \cite{per}, which reweights how often stored transitions are sampled but never discards them as obsolete. It is instead the deep-world-model counterpart of tempering a prior on a changepoint \cite{arcade}: keep what generalizes (the representation), discard what
is now wrong (the data).

\textbf{Respawn baseline:} As a controlled alternative we replace the forgetting step
with an MBCD-style \emph{respawn}: on detection, reinitialize the world model's latent
dynamics and prediction heads in place (a fresh ``context model''), keeping the encoder
and policy, and reset the model optimizer. We emphasize this is an MBCD-\emph{style}
response adapted to a deep world model, not a literal reimplementation of MBCD's
ensemble-and-model-bank system. It isolates the single question of whether to
\emph{discard the model} or merely \emph{discard the data} on a shift.

\section{Experiments}
\textbf{Setup:} We use the \texttt{walker\_walk} locomotion task from the DeepMind
Control suite \cite{dmc} on the MuJoCo physics engine \cite{mujoco}, with a DreamerV3
agent. We inject two
robot-relevant abrupt shifts: gravity $\times 2.0$ (a payload / load-change analogue)
and actuator-gain $\times 0.5$ (a motor-wear analogue), each fired at a fixed frame. To
compare responses on identical histories we train one pre-shift checkpoint to
convergence and branch three arms that resume it and undergo the shift: \emph{passive}
(continue training, full buffer), \emph{CAWM} (forget on the shift), and
\emph{MBCD-style} (respawn on the shift). To isolate the response from detector-timing
noise, the three-arm comparisons below trigger the response at the known shift frame
(oracle timing). We then close the loop with the online detector and report it
separately, verifying it fires within a few hundred frames of the shift with no
pre-shift false alarm. We report mean $\pm$ std over three seeds.

\begin{figure*}[t]
\centering
\includegraphics[width=0.82\textwidth]{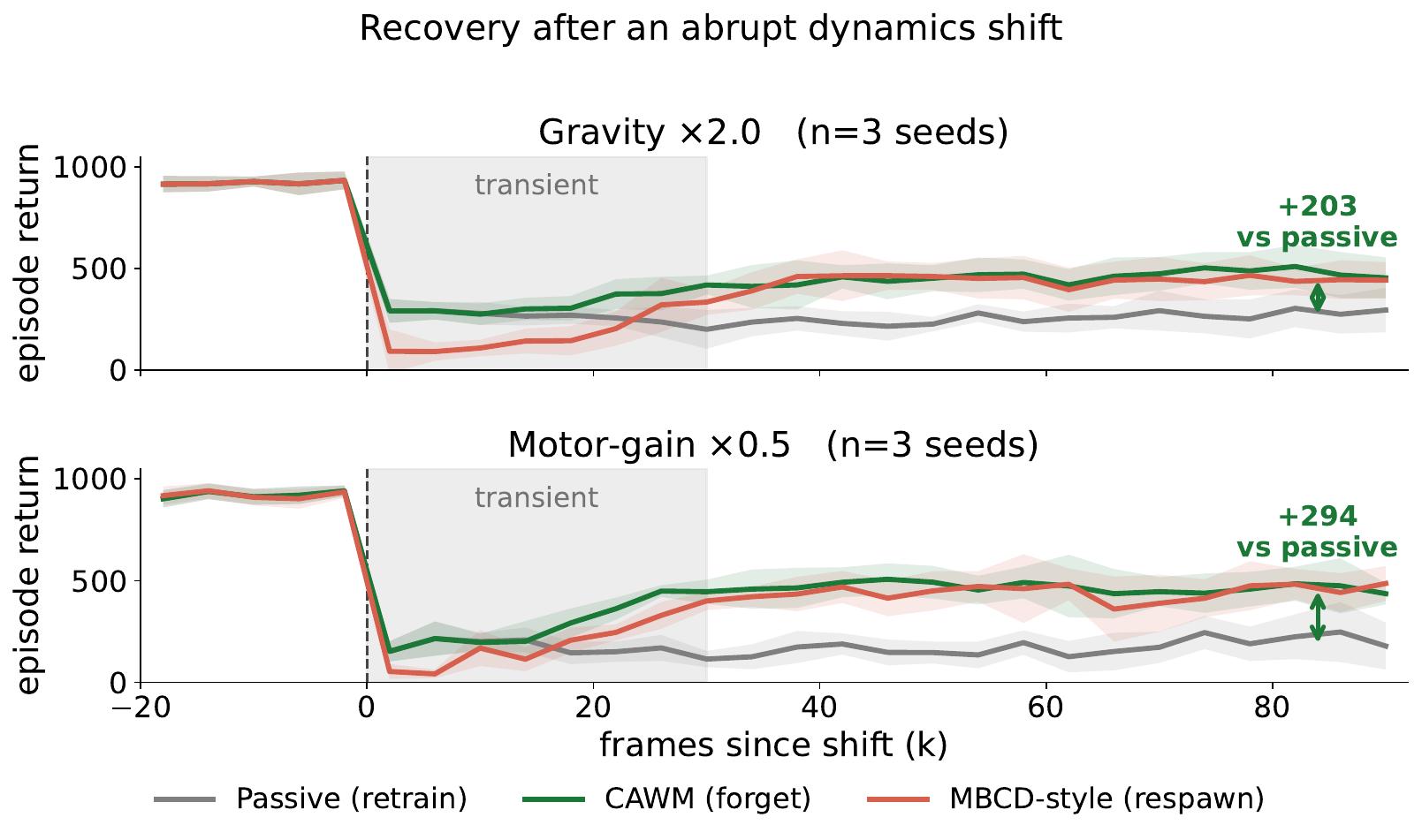}
\caption{Episode return versus frames since the shift, for the gravity (top) and
motor-wear (bottom) shifts, mean over three seeds (shaded: std), with the transient
window highlighted. Passive retraining stays stuck, CAWM recovers fastest, and the
MBCD-style respawn crashes hardest immediately after the shift as its fresh model
relearns before catching up. The green arrow marks CAWM's asymptotic gain over passive.}
\label{fig:recovery}
\end{figure*}

\textbf{Recovery.} Figure~\ref{fig:recovery} shows the dynamics. Under both shifts,
passive retraining never recovers, pulled down by its stale buffer. CAWM recovers
quickly and to the highest level. The MBCD-style respawn suffers a second, sharp
crash right after the shift, the cost of relearning a dynamics model from scratch,
before re-climbing to roughly CAWM's level. Table~\ref{tab:results} quantifies this
over the three windows. CAWM beats passive decisively everywhere (up to $+294$ at
asymptote under motor-wear). Against the MBCD-style respawn, CAWM's advantage is
concentrated in the \emph{transient}: $+153$ on gravity and $+95$ on motor-wear in the
first $30$k frames, robust across all three seeds. CAWM never pays the respawn's
re-adaptation crash. At \emph{asymptote} the two are comparable ($+18$ and $+27$, within
seed variation), so forgetting recovers faster, not higher. The cumulative window, a
return-based regret proxy, favors CAWM by $+63$ and $+50$.

\begin{table}[t]
\caption{Mean post-shift return (\texttt{walker\_walk}, three seeds). Windows are
frames since the shift. The $\Delta$ values discussed in the text are CAWM minus each baseline.}
\label{tab:results}
\centering
\begin{tabular}{@{}llrrr@{}}
\toprule
Shift & Window & Passive & MBCD-style & CAWM \\
\midrule
\multirow{3}{*}{Gravity $\times2.0$}
 & Transient  & $268$ & $171$ & $\mathbf{323}$ \\
 & Asymptote  & $256$ & $441$ & $\mathbf{459}$ \\
 & Cumulative & $260$ & $351$ & $\mathbf{414}$ \\
\midrule
\multirow{3}{*}{Motor-gain $\times0.5$}
 & Transient  & $172$ & $183$ & $\mathbf{278}$ \\
 & Asymptote  & $173$ & $440$ & $\mathbf{467}$ \\
 & Cumulative & $173$ & $354$ & $\mathbf{404}$ \\
\bottomrule
\end{tabular}
\end{table}

\begin{figure*}[t]
\centering
\includegraphics[width=0.82\textwidth]{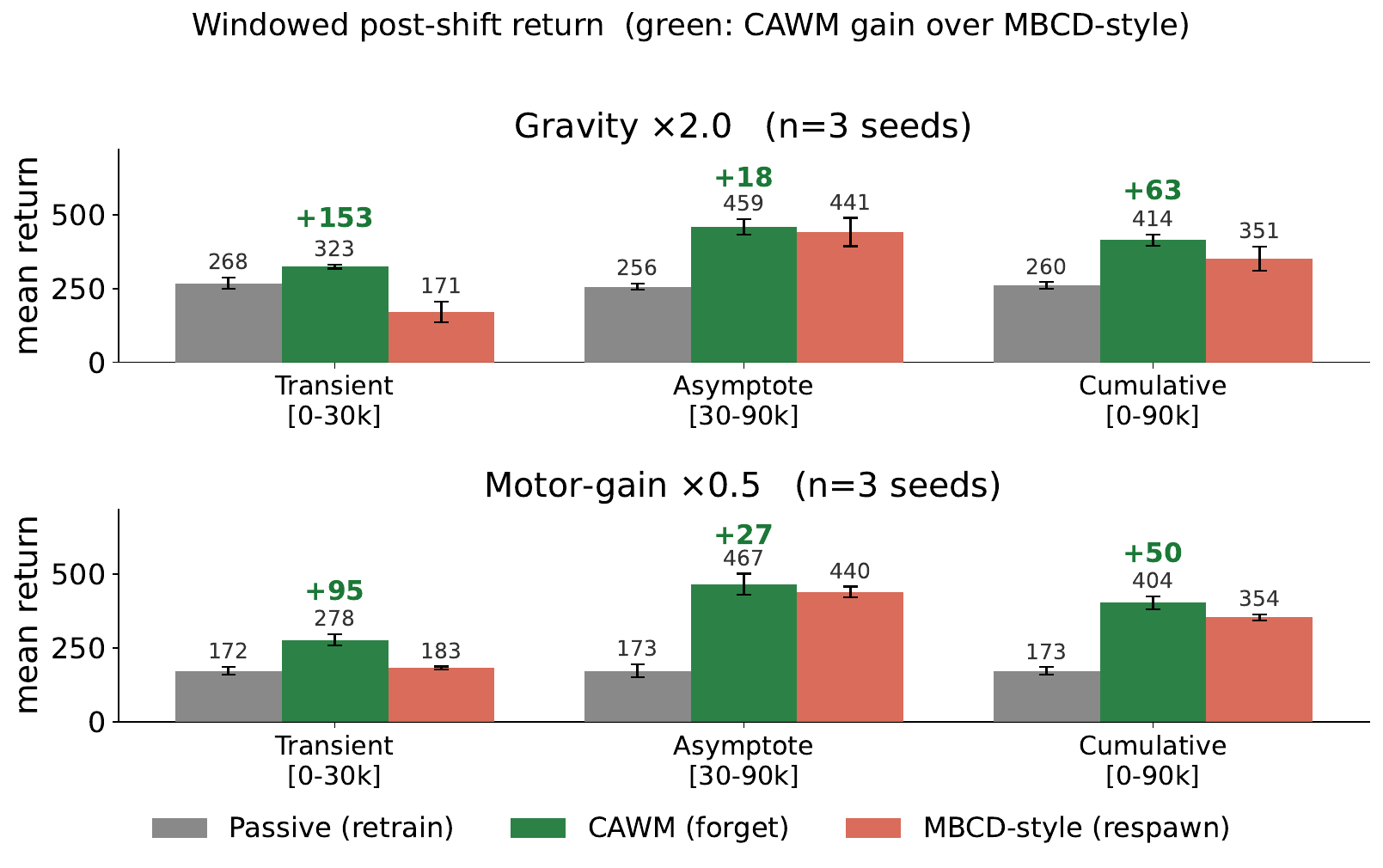}
\caption{Windowed mean return for the gravity (top) and motor-wear (bottom) shifts,
three seeds (error bars: std). Green annotations give CAWM's gain over the MBCD-style
respawn: large in the transient, shrinking to a tie at asymptote, on both shifts.}
\label{fig:windows}
\end{figure*}

\textbf{Consistency across shift types.} The two shifts are physically unrelated, a
global force change versus a per-actuator gain change, yet produce the same structure
(Table~\ref{tab:results}, Fig.~\ref{fig:windows}). The same detect-and-forget mechanism holds across both shifts without per-shift tuning, evidence that the effect is about \emph{the response to non-stationarity}, not a particular perturbation.

\textbf{Closed-loop detection.} The comparisons above trigger the response at the known
shift frame to isolate the response mechanism from detector-timing noise. We also ran
the full loop with the online detector in control. On the gravity $\times 2.0$ shift,
the CUSUM test fires $200$ to $1000$ frames after the true changepoint across three
independent seeds, with \emph{no} pre-shift false alarm despite the agent's ongoing
learning drift. That drift is the precise failure mode a fixed baseline suffers and the
rolling baseline removes. The detector-triggered forgetting then recovers to the same
level as the oracle-timed response, beating passive by $+203\pm18$ over the asymptote
window. Because the detection delay and the buffer-flush lag are small
relative to the multi-thousand-frame window over which passive stays broken, closing
the loop costs little: the recovery benefit is preserved end-to-end with essentially no
added machinery.

\textbf{Robustness across seeds.} The transient advantage over the respawn is the
consistent part of the result: CAWM wins the first-$30$k window in all three seeds of
both shifts. The asymptotic margin behaves differently, on motor-wear it shrinks across
seeds ($+56$, $+23$, $+2$), so the two arms converge and their per-seed ordering there
is not stable. We therefore report the asymptote as a tie and claim only faster
recovery, which the data robustly support. The CAWM-over-passive gap is large and stable in every window and seed.

\textbf{When does forgetting help?} Forgetting is not universally good. Under a milder
gravity shift ($\times 1.5$) the passive agent self-heals to roughly its pre-shift
return ($\sim$$900$) because much of its old experience is still valid, and forgetting
then offers no benefit, plateauing lower ($\sim$$716$) as it discards data that is still
useful. Forgetting helps precisely when the shift is severe enough that pre-shift data is
obsolete (gravity $\times 2.0$, gain $\times 0.5$), where passive retraining is
permanently dragged down (Table~\ref{tab:results}). Whether to forget is thus
shift-magnitude dependent, which the changepoint statistic's magnitude can inform but a
fixed ``always forget'' rule ignores.

\textbf{Implementation details.} The base agent is DreamerV3 in its default
proprioceptive configuration (action repeat $2$). A single pre-shift checkpoint is
trained for $150$k frames to convergence and saved with its replay buffer. All three
arms resume it and run $130$k further frames, of which the last $90$k are post-shift.
Forgetting caps the replay buffer from its pre-shift size (${\sim}1.5{\times}10^5$
transitions) to $N_{\mathrm{keep}}{=}10^4$, after which DreamerV3's existing
oldest-first eviction flushes stale data within ${\sim}N_{\mathrm{keep}}$ frames. The
detector consumes one $\mathcal{L}_{\mathrm{dyn}}$ value per logging step (every $100$
frames) with CUSUM slack $k{=}0.5$, threshold $h{=}8$, a trailing baseline of the last
$100$ values (lag $5$), light exponential smoothing, and a $60$-step warm-up before it
may fire. It thus adds one scalar test per logging step and no extra network
evaluations. The respawn
baseline reinitializes the latent dynamics and prediction heads and resets the model
optimizer, keeping the encoder and actor-critic. Forgetting and respawn thus differ only
in whether the learned dynamics are discarded.

\textbf{Scope.} The detector assumes a roughly converged pre-shift baseline, matching an
agent that has settled before its dynamics change. The seed arms share that pre-shift
checkpoint and vary post-shift stochasticity. Broadening to more tasks, gradual drift,
and fully independent seeds is future work.

\section{Discussion}
These results suggest that, for a model-based agent facing abrupt dynamics change, the
leverage is in \emph{what to discard}. Discarding the data while keeping the
representation (CAWM) recovers faster than discarding the model (respawn) and far faster
than discarding nothing (passive): the learned encoder still generalizes across the
shift, so only the dynamics need re-fitting. An abrupt shift changes how states evolve,
not how observations map to states. That is why the representation survives while the
transition model alone must be relearned. The advantage is a recovery \emph{transient},
not an asymptotic ceiling, and it appears only when the shift is severe enough to render
old data obsolete. The detector also adds little cost. It reuses a loss the world model computes and introduces no extra networks, so the mechanism drops into any Dreamer-style agent
without changing the base learner. The most immediate next direction is breadth across robot morphologies, extending the same detect-and-forget comparison to the cheetah and quadruped bodies to test whether the transient advantage holds as the dynamics grow higher-dimensional. The conditional result also motivates graded forgetting scaled to the changepoint magnitude rather than a fixed cap, stronger non-stationary baselines such as a recency-weighted buffer and an explicit model bank, and calibrating the detector to a target false-alarm rate.

\section{Conclusion}
We presented Changepoint-Aware World Models, which detect an abrupt dynamics shift from
a deep world model's own prediction error and recover by forgetting stale replay while
keeping the representation. Across two robot-relevant shifts and three seeds, this
recovers faster than both passive retraining and respawning a fresh dynamics model, at
comparable asymptote.

\bibliographystyle{plainnat}
\bibliography{references}

\end{document}